\documentclass[twoside]{article}
\usepackage{PRIMEarxiv}

\usepackage{times}
\usepackage{soul}
\usepackage{url}
\usepackage[hidelinks]{hyperref}
\usepackage[utf8]{inputenc}
\usepackage[small]{caption}
\usepackage{graphicx}
\usepackage{subcaption}
\usepackage{amsmath}
\usepackage{amsthm}
\usepackage{booktabs}
\usepackage{algorithm}
\usepackage{algorithmic}
\usepackage[switch]{lineno}
\usepackage{amssymb}
\usepackage{booktabs}
\usepackage[dvipsnames]{xcolor}
\usepackage{svg}

\usepackage[dvipsnames]{xcolor}
\newcommand{\ngram}{n-gram}

\title{Intertwining CP and NLP: \\
The Generation of Unreasonably Constrained Sentences}

\author{
Alexandre Bonlarron \\
Université Côte d’Azur, Inria, France \\
Université Côte d’Azur, CNRS, I3S, France \\
alexandre.bonlarron@gmail.com
\and
Jean-Charles Régin \\
    Université Côte d’Azur, CNRS, I3S, France\\
    jcregin@gmail.com
}

\begin{document}
 \maketitle
\begin{abstract}
   Constrained text generation remains a challenging task, particularly when dealing with hard constraints. Traditional NLP approaches prioritize generating meaningful and coherent output. Also, the current state-of-the-art methods often lack the expressiveness and constraint satisfaction capabilities to handle such tasks effectively. Recently, an approach for generating constrained sentences in CP has been proposed in \cite{bonlarron-et-al:2023}. This ad-hoc model to solve the sentences generation problem under MNREAD rules proved neithertheless to be computationaly and structuraly unsuitable to deal with other more constrained problems. In this paper, a novel more generic approach is introduced to tackle many of these previously untractable problems, and illustrated here with the quite untractable sentences generation problem following RADNER rules.

More precisely, this paper presents the \emph{CPTextGen} Framework. This framework considers a constrained text generation problem as a discrete combinatorial optimization problem. It is solved by a constraint programming method that combines linguistic properties (e.g., n-grams or language level) with other more classical constraints (e.g., the number of characters, syllables). Eventually, a curation phase allows for selecting the best-generated sentences according to perplexity using an LLM.

The effectiveness of this approach is demonstrated by tackling a new, more tediously constrained text generation problem: the iconic RADNER sentences problem. This problem aims to generate sentences respecting a set of quite strict rules defined by their use in vision and clinical research. Thanks to our CP-based approach, many new strongly constrained sentences have been successfully generated. This highlights our approach's potential to handle unreasonably constrained text generation scenarios.
\end{abstract}

\section{Introduction}

Constrained text generation is a great challenge in natural language processing (NLP), demanding the creation of textual output that adheres to specific requirements. This task encompasses a broad spectrum of text generation scenarios, ranging from crafting lyrics adhering to rhythmic, lexical, and metrical constraints to producing text summaries prioritizing length, semantic entailment, or other desired properties.
This article focuses on the first category of constrained output, where strict adherence to specific requirements is paramount. 

In NLP, several methods have been developed to tackle contained text generation. The major approach relies on a Large Language Model (LLM) guided search \cite{lu-etal-2022-neurologic,beamsearch1:2021,beamsearch2:2018,hokamp-liu:2017}.
For instance, BeamSearch excels when there are lexical constraints (i.e., enforce or avoid a set of keywords). 

In addition, the latest series of conversational agents like ChatGPT \cite{openai2023gpt4}, PaLM \cite{anil2023palm}, Alpaca, Vicuna \cite{touvron2023llama} show that LLM can achieve decent constraint satisfaction rates (99\%) on lexical constrained generation task (see sent-04 task in \cite{yao2023collie} for more information). Nevertheless, at the same time, even fine-tuned prompted LLMs fail to generate a sentence that contains for example exactly 82 characters (see sent-01 task).

Intuitively, enforcing that a sentence must contain exactly 82 characters can be reduced to a knapsack problem where each word has a weight equal to its number of characters and where the total number of characters of the sentence is the capacity of the backpack (i.e., 82). It is known to be an NP-complete problem. 

In the NLP taxonomy, this requirement type is called hard constraint \cite{garbacea-arxiv-survey-nlp:2022}, characterized by binary functions associated with the constraints that must be true. Interestingly, the concept of hard constraints in NLP aligns with the classical definition of a constraint in discrete combinatorial optimization (CO). Various hard constraints in NLP find their counterparts in CO, such as the length constraint in NLP corresponding to a knapsack constraint in CO. In addition, the knapsack constraint has been well-studied in CO as it is ubiquitous in real-world problems. Moreover, several efficient algorithms exist to solve this constraint using dynamic programming \cite{trick:03}.
Thus, using a dynamic programming approach to compute a set of sentences of 82 characters sounds promising.

Furthermore, over the past decade, research in assisted song composition, spearheaded by Pachet's ERC Flow-Machines\footnote{https://www.flow-machines.com}, has highlighted the connection between output specifications in songs or texts and constraints in the context of CO. Various output requirements were successfully modeled and solved in CO, including style modeling \cite{pachet-roy:2011}, virtuoso melodies \cite{pachet-roy-barbieri:2011-virtuoso}, lyrics \cite{barbieri-et-al-lyrics:2012}, meter \cite{Roy-Pachet:2013-meter}, plagiarism \cite{papadopoulos-roy-pachet:2014}, palindrome \cite{papadopoulos-roy-etal:15}, poetry \cite{perez-regin:17b} and
standardized sentences \cite{bonlarron-et-al:2023}. Notably, this research has paved the way for formulating constrained generation tasks in text and music as constraint satisfaction problems (CSPs).

In the context of text generation under constraints, viewing this problem as a CSP implies defining: (1) The set of decision variables that represent words of a sentence. (2) The domain of each variable is the set of allowed vocabulary. (3) Finally, the constraints are the set of rules that the text (the assigned variables) must satisfy.

A well-known technique to solve a CO problem when formalized as a CSP is Constraint Programming (CP). CP is a paradigm that computes solutions that satisfy constraints. CP is mainly based on filtering algorithms (also known as propagators), which remove values from variable domains that do not belong to a solution of a constraint. The association of such algorithms with global constraints is one of the main strengths of CP because they exploit the specific structure of each constraint.
Thus, CP is an excellent candidate for tackling constrained generation tasks.

In this paper, we introduce \emph{CPTextGen} a new CP-based framework to generate sentences under constraints that overcomes the limitations of the ”Constraints
First” approach described in \cite{bonlarron-et-al:2023}.
This framework combines the semantic aspects of sentences in a given language with the constraints imposed on the words that make them up.
On the one hand, it uses n-grams and possibly their associated likelihoods. On the other hand, it uses variables whose domains are words and constraints between variables, to allow, for example, a limit on the number of characters used.
The framework proposes to transform the initial problem into a combinatorial optimization problem based on the multi-valued decision diagram (MDD) data structure, which is particularly well-suited to representing a corpus. This problem is then solved using constraint programming. 

Additionally, it introduces the problem of RADNER sentence generation, which is a clinical test for people with potential vision problems. This problem is used to demonstrate the interest of our framework in solving a particularly difficult problem for which no generic solution is known to date. 

The structure of this paper is outlined as follows:
\begin{itemize}
    \item  Section~\ref{sec:Prelim}: Background information on Constraint Programming (CP) and NLP is introduced.
    \item Section~\ref{sec:method}: A constraint programming model is presented for generating constrained sentences. 
    \item Section~\ref{sec:case-study}: The RADNER sentence problem is formalized as a CSP.
    \item  Section~\ref{sec:resultats}: The effectiveness of the proposed approach is showcased by detailing the generated output.
    \item  Section~\ref{sec:discussion}: Open challenges, limitations, and future perspectives associated with the approach are discussed.
    
\end{itemize}

\section{Preliminaries}
\label{sec:Prelim}
\subsection{Multi-valued Decision Diagram}

Multi-valued decision diagrams (MDDs) are directed acyclic graphs (DAG) data structures used in discrete optimization. They consist of nodes and arcs, with two particular nodes: the root node and the terminal node. Here, only Deterministic Reduced Ordered MDDs \cite{Amilhastre:2014} are considered.

MDDs are mainly used as archives for \emph{Table Constraint}~\cite{cheng-yap:10,lecoutre:2011} and are generic to compute and store various other constraints~\cite{Malalel-et-al:2023,Perez_Malalel_Glorian_Jung_Papadopoulos_Pelleau_Suijlen_Régin_Lallouet_2023,Wang-Yap:2022,gentzel-etal:2022,verhaeghe2019extending,verhaeghe2018compact}. MDDs as a tool for solving optimization problems can be found in ~\cite{,gillard2022large,rudich:2022peelandbound,Hoeve:2022,bergman-cire-etal:16}. 
\color{black}

MDDs are organized into layers, where each layer contains nodes and outgoing arcs. Each layer represents a variable, and therefore, any MDD model represents an n-ary function $f(x_1, x_2, ..., x_n) \mapsto \{true, false\}$. Equivalent nodes (i.e., nodes with the same outgoing arcs with the same labels) are merged, compressing the data structure. Various operations (intersection, union, difference...) can be applied between MDDs without the need to decompress them.

\subsubsection{MDD of a Constraint}

Given a constraint $C$ defined over a set of variables $X = \{x_1, x_2, ..., x_n\}$, an MDD can be compiled to store the valid assignments (tuples) associated with the constraint $C$. Therefore, MDD$(C)$ is the MDD associated with the constraint $C$.

\subsubsection{Cost-MDD}
A cost-MDD is an MDD whose arcs have additional information: the cost $c$ of the arc. Let $M$ be a cost-MDD and $p$ be a path of $M$. The cost of $p$ is denoted by $\gamma(p)$ and is equal to the sum of the costs of the arcs it contains.

\subsection{Constraint Satisfaction Problem}

A Constraint Satisfaction Problem (CSP) is a triplet: $\langle \mathcal{X}, \mathcal{D}, \mathcal{C} \rangle$, where:
\begin{itemize}
\item $\mathcal{X} = \{X_1, X_2, ..., X_n\}$ is the set of variables  of the problem.
\item $\mathcal{D} = \{D_{X_1}, D_{X_2}, ..., D_{X_n}\}$ is the set of domains, where each domain $D_{X_i}$ corresponds to the set of possible values for the variable $X_i$.
\item $\mathcal{C} = \{C_1, C_2, ..., C_m\}$ is the set of constraints of the problem. A constraint represents a property of the solutions to the problem.
\end{itemize}
% A solution to a CSP 
A solution is an assignment of all the variables to a value present in their respective domains, such that all the constraints are satisfied.

\begin{figure}[htbp]
  \begin{center}
    \includegraphics[width=0.33\textwidth]{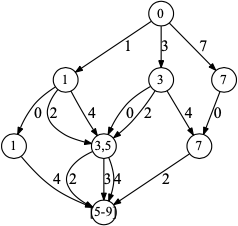}
  \end{center}
\caption{\label{fig:exemple-MDD-type-somme}
Example of MDD representing the set of %solutions 
$x_1 + x_2 + x_3 \in [5,9]$.
For each variable $x_i$, the domain $D(.)$ is %well defined:
$D(x_1)$ = $\{1,3,7\}$, $D(x_2)$ = $\{0,2,4\}$, $D(x_3)$ = $\{2,3,4\}$.
For example, $(7,0,2)$ belongs to the set of solutions defined by the MDD. }
\end{figure}

\subsection{Language Model}
\subsubsection{N-gram Model}
An n-gram \cite{shannon-ngram:51,jurafsky:2009} is a contiguous sequence of $n$ words extracted from a text corpus. Consider the sentence, ``The princess wears a red dress". This sentence contains several 3-grams, such as ``The princess wears," ``princess wears a," ``wears a red," and ``a red dress" We can define chaining rules between n-grams based on shared word n-tuples. Two n-grams are considered chained if they share a common sequence of $n-1$ consecutive words. More precisely, the last $n-1$ words of the first n-gram must be identical to the first $n-1$ words of the second n-gram.
\subsubsection{LLM}
Recent advancements in NLP have witnessed the emergence of powerful language models (LLMs) (e.g., PaLM \cite{anil2023palm}, GPT-4 \cite{openai2023gpt4}, LLaMa \cite{touvron2023llama}) leveraging the Transformer architecture \cite{vasmani-et-al:2017}. These models are trained on massive datasets, enabling them to generate human-quality text through various decoding strategies, such as greedy decoding or more complex strategies. As language models, their core function relies on estimating the probability of a given word sequence.

\subsubsection{Perplexity}
Text perplexity (PPL), a measurement of uncertainty rooted in Shannon's information theory, can be readily computed in large language models (LLMs), given their inherent ability to calculate text probabilities. It represents the geometric mean of the inverse conditional likelihoods associated with a given sequence of words \cite{jurafsky:2009}. Formally, let $S_n$ denote a sequence of $n$ words, written as $S_n = w_1w_2\dotsm w_n$. The PPL of $S_n$ is calculated as:

 $$PPL(S_n) = \sqrt[n]{\frac{1}{P(w_1w_2\dotsm w_n)}}$$

 where $P(\cdot)$ denotes the LLM's assigned probability. Intuitively, PPL reflects the model's perceived likelihood of generating a particular text, with lower values signifying higher confidence \cite{garbacea-arxiv-survey-nlp:2022}. This metric is valuable for LLM evaluation, ensuring that the model correctly identifies well-formed samples through low PPL scores.
\section{The CPTextGen Framework}
\label{sec:method}
                                                 
\begin{figure}[tbp]
    \centering
    \resizebox{0.5\textwidth}{!}{ 
    \includegraphics{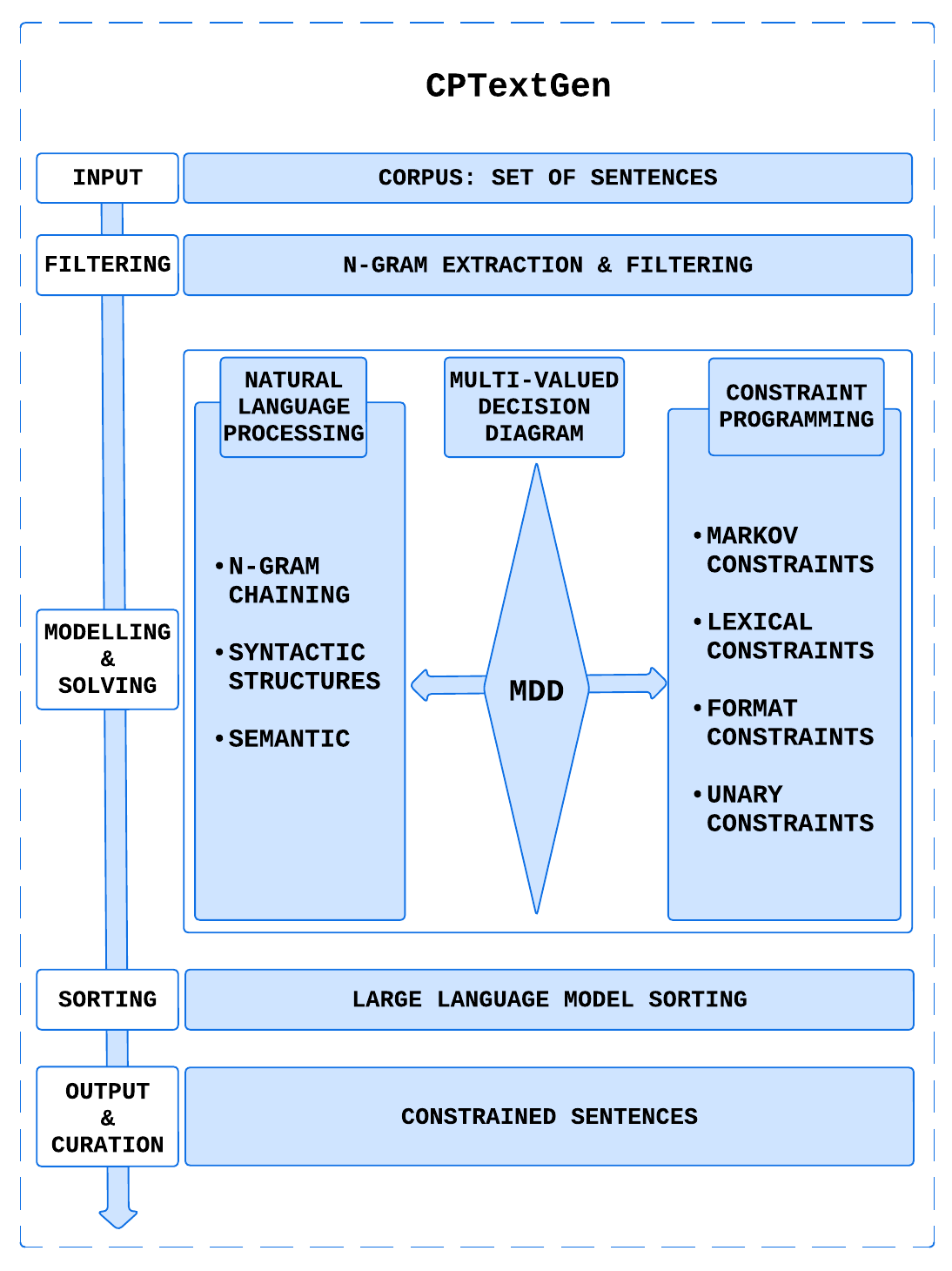}
    }
    \caption{This figures summarizes the major step of the \emph{CPTextGen} framework. Once the filtered n-grams are gathered. The MDD data structure acts as a bridge between the constraint satisfaction relying on CP techniques and the n-gram chaining that takes account of the structure of the language.}
    \label{fig:ConstraintFirst}
\end{figure}

This section presents the major steps of the \emph{CPTextGen} framework. An overview is depicted in Fig. \ref{fig:ConstraintFirst}.

\subsection{Input Data: N-gram Corpora}
First, a corpus (i.e., a set of sentences) is needed. This corpus can be made from various sources (e.g., books, newspapers, subtitles, Wikipedia,...). From the sentences set, all n-grams are extracted, and some are filtered out if necessary. For instance, if some words are forbidden, all n-grams where the words appear are deleted.

Using n-grams is essential for integrating language into generation. A sentence in a language is very different from a random sequence of words. Although it is difficult to manage the meaning of a sentence (i.e., generate only sentences that make sense), this does not mean that the meaning of a sentence should be completely abandoned during generation.

\subsection{Ngram reTRIEval}
\begin{figure}[tbp]
    \centering
    \includesvg[width=0.41\textwidth]{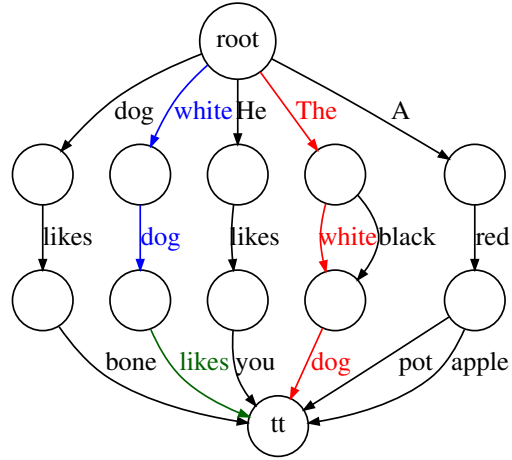}
    \caption{Example of MDDtrie storing 3-grams (successions of 3 words): ``The black dog"; ``A red pot"...
    Any path from the \emph{root} to \emph{tt} is a valid \ngram. 
    To find the successors of the \ngram~  ``{\color{red}The white dog}" (in red), more precisely the following potential words,
 we start from \emph{root} to walk along the arcs that contains the labels of the two last arc, i.e., ``{\color{blue}white}" and ``{\color{blue}dog}" (in blue). In that case, one outgoing arc from the node can be reached with ``white cat". Thus, the successor of ``The white dog" is ``{\color{OliveGreen}likes}" (in green).}

    \label{fig:MDDtrie}
\end{figure}

Next, the set of n-grams is stored within an MDD (see Fig.~\ref{fig:MDDtrie}). 
This MDD is used as a Trie; thus, it is named MDDTrie. It manages the operations to store and retrieve n-grams.

\subsection{Constraint Programming Model}

The Constraint Programming Model is based on an MDD that was built from MDDTrie.
It contains all the feasible sentences satisfying the constraints.

\begin{table}[tbp]
    \centering
    \resizebox{0.66\textwidth}{!}{  
    \begin{tabular}{ll}\toprule
    
     & A Constraint Programming Model of Standardized Sentences \\
      \midrule
     $\mathcal{V}$ & $X_1,X_2,...,X_{N}$ \\
      $\mathcal{D}$ & $\forall i, D(X_i) = \{words \in AllowedLexicon\}$ \\
        \midrule
         & \multicolumn{1}{c}{Language modelling with k-grams} \\
        \midrule
        $C_{LM^{start}}$ & $IsNgramOfStart(X_1,X_2,X_3,...,X_k) = \top$ \\
        $C_{LM^{succ}}$ & $i>k$, $X_i = NgramSuccession(X_{i-k},...,X_{i-3},X_{i-2},X_{i-1})$ \\
        $C_{LM^{end}}$ & $IsNgramOfEnd(X_{14-k},X_{14-k-1},...,X_{14}) = \top $\\
        
       \midrule
        $C_{Length}$& \multicolumn{1}{c}{Format (length) Constraints} \\
       \midrule 
       $C_{L^1}$ &  $\lfloor LineSize \rfloor \leq \sum_{i=1}^{L_1} \#Char(X_i) \leq \lceil LineSize \rceil$  \\ 
       \addlinespace[0.25em]
        $C_{L^2}$ &  $\lfloor LineSize \rfloor \leq \sum_{i=1+L_1}^{L_2} \#Char(X_i) \leq \lceil LineSize \rceil$  \\
        \addlinespace[0.25em]
        $C_{L^3}$ &  $\lfloor LineSize \rfloor \leq \sum_{i=1+L_2}^{L_3} \#Char(X_i) \leq \lceil LineSize \rceil$  \\ 
        \addlinespace[0.25em]
         $C_{L^4}$& $\lfloor SentenceSize \rfloor \leq \sum_{i=1}^{N} \#Char(X_i) \leq \lceil SentenceSize \rceil$ \\
         \addlinespace[0.25em]
         \midrule
        $ C_{Meter}$& \multicolumn{1}{c}{Meter (syllable) Constraints} \\
         \midrule
          $C_{M^1}$ &  $\sum_{i=1}^{N} \#Syll(X_i) = SyllableCount$ \\
   
         \bottomrule
    \end{tabular}
   }
   \caption{Minimalistic CSP model to generate generic standardized sentences. The decision Variables are words, and the Domains are the authorized words. Then, the n-gram constraints ensure that any assignment of the variables is consistent with the n-gram chaining rules.
   Finally, several knapsack constraints are defined to control the total size of the sentence, its display property (3 lines, for instance), or its meter property.
   $\#Syll(X_i)$ is the number of syllables of the assigned variable $X_i$ and $\#Char(X_i)$ is the number of characters of $X_i$. }
   \label{CP:Model}
\end{table}

\subsubsection{MDD Compilation}

MDD can be used to encode and solve CSP \cite{hoda-et-al:2010,perez-regin:15}. Figure \ref{fig:exemple-MDD-type-somme} gives an example of such encoding. In particular, an MDD-based model has been shown to be nearly adequate for this task \cite{bonlarron-et-al:2023}. The approach developed claimed to be general, but its description is too close the resolution of the MNREAD sentences problem. 

The CSP model is compiled into MDD (i.e., each constraint in the CP model (see Tab. \ref{CP:Model}) is associated with a particular MDD). The number of layers of the MDD is the number of variables. The potential label that an arc can have is a function of the domains of the associated variables of the layer. The number of characters or syllables of a word is a cost. Two cost functions, respectively, $\#Char(.)$ and $\#Syll(.)$, are available for the label of an arc. Thus, all MDDs in this approach are cost-MDDs.

The MDD $MDD_{final}$ containing all the solutions that satisfy the conjunction of the various constraints can be obtained in two ways. Either there is an interest in keeping intermediate results, and $MDD_{final}$ is computed by intersecting the MDDs associated with each explicitly defined constraint. Alternatively, the intersection of the MDDs associated with each constraint is performed on the fly. This means that when performing an intersection between $MDD_1$ and $MDD_2$ s.t., $MDD_1~\bigcap~MDD_2$, $MDD_2$ is given in intention, so its solutions are never computed. This second approach saves a lot of memory and is faster. Formally, we have

\[ MDD_{final} = \bigcap^{m}_{i=1}{MDD(C_i)}  \]
therefore,
\[ MDD_{final} = MDD(\wedge^{m}_{i=1}{C_i})\]

So far, we have assumed that it is possible to build $MDD_{final}$ explicitly by successive intersections, whatever constraints are added. This turns out to be the case for our application domain. This approach is quite realistic for the types of constraints we have been led to integrate. However, should it become impossible to calculate MDD intersections for new constraints, this does not render the proposed framework useless. Indeed, there is no need to build the final MDD using intersections alone. We could very well build an MDD integrating a subset of the constraints, then use a CP solver integrating this MDD (using the MDD4R propagator) with the constraints not yet integrated to calculate the solution set of the problem.

\subsection{Sentence Curation Helped by an LLM}
Once the $MDD_{final}$ is computed, all solutions (sentences) can be extracted. The solution set is then evaluated by LLM. Thus, the solutions are sorted by the perplexity score computed by an LLM. This simplifies the sentence selection process dramatically.

\section{Case Study: Standardized Text}
\label{sec:case-study}
\subsection{Why Standardized Text Is Critical?}

Reading performance assessment has become a prominent method for evaluating vision \cite{radner:17}. Standardized text, particularly sentences, has established itself as a cornerstone for reading performance assessments in vision research \cite{mansfield-atilgan-etal:19}. Due to the intricate nature of reading involving various individual-specific parameters, standardized materials effectively minimize person-related biases in measurements \cite{legge2016reading}. Several reading charts utilizing standardized sentences exist, such as the MNREAD test \cite{mansfield-ahn-etal:93} or the RADNER test \cite{radner-english:2014}. While sharing a common purpose, each chart possesses its strengths and limitations. Vision research necessitates standardized sentences and tools that empower researchers to readily produce standardized material to assess reading performances and test hypotheses related to visual impairment. 

This motivation originating from the vision research field converges with a compelling challenge in artificial intelligence regarding the state-of-the-art in constrained text generation. To the best of our knowledge, all existing RADNER sentences were handcrafted. Therefore, this paper introduces the first approach, which is able to generate them automatically. Solving this real-world problem leads to thinking about the potential intertwining of CP and NLP.

Our new \emph{CPTextGen} framework is quite comprehensive and incorporates, for example, but not only, previously developed approaches such as the "Constraints First" model developed in \cite{bonlarron-et-al:2023} to solve the MNREAD sentence problem. However, in order to illustrate the potential of our framework, in this paper we introduce a completely new model within the framework to solve another more challenging sentence generation problem, the RADNER sentences problem.

\subsection{The RADNER Sentences Problem}

\begin{figure}[tbp]
    \centering
    \includegraphics[width=\textwidth]{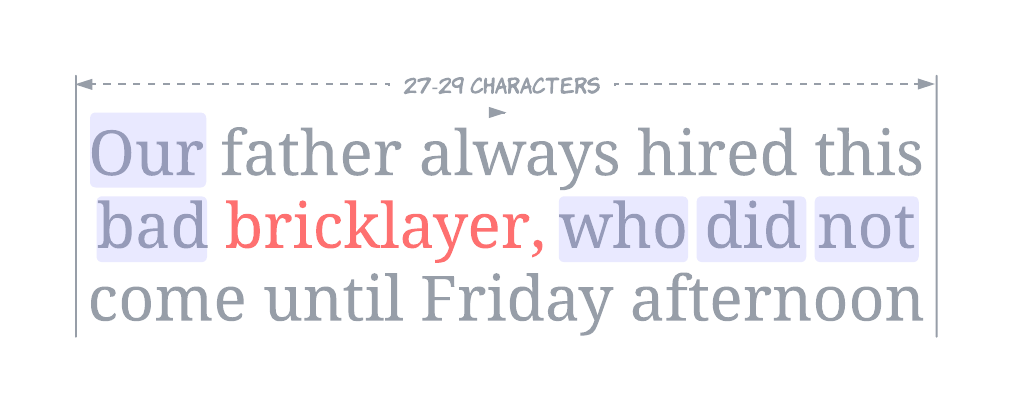}
    \caption{Example of an English RADNER sentence incorporating relative clauses with three lines and 14 words.
    A subset of the RADNER rules are highlighted.
    Each line contains between 27 and 29 characters; also, in purple, several words should be of exactly one syllable.
    In red, the second word of the second line must contain 10 characters and three syllables. }
    \label{fig:enter-label}
\end{figure}

The basic rules that a RADNER sentence must obey are the following : 
\subsubsection{RADNER Core Rules}
\begin{itemize}
    \item (1) The sentences incorporated relative clauses and had three lines and 14 words.
    \item (2) Each line had 27–29 characters, including spaces (82–84 characters per sentence).
    \item (3) The first and second lines had 5 words.
    \item (4) The third line had 4 words.
\end{itemize}

\begin{table}[t]
    \centering
    \resizebox{0.66\textwidth}{!}{  
    \begin{tabular}{ll}\toprule
    
     & A Constraint Programming Model of Radner-like Sentences \\
     \midrule
    $\mathcal{V}$ & $X_1,X_2,...,X_{15}$ \\
     $\mathcal{D}$ & $\forall i, D(X_i) = \{words \in Vocabulary\}$  \\
        \midrule
         & \multicolumn{1}{c}{Language Modelling Constraints} \\
        \midrule
        $C_{LM^{start}}$& $IsNgramOfStart(X_1,X_2,X_3,...,X_k) = \top$\\
        $C_{LM^{succ}}$& $i>k, X_i = NgramSuccession(X_{i-k},...,X_{i-3},X_{i-2},X_{i-1})$ \\
        $C_{LM^{end}}$& $IsNgramOfEnd(X_{14-k},X_{14-k-1},...,X_{15}) = \top$\\
        
       \midrule
        $C_{Length}$& \multicolumn{1}{c}{Format (length) Constraints} \\
       \midrule 
       $C_{L^1}$ & $ 27 \leq \sum_{i=1}^{5} \#Char(X_i) \leq 29 $ \\ 
       \addlinespace[0.25em]
        $C_{L^2}$ & $27 \leq \sum_{i=6}^{11} \#Char(X_i) \leq 29 $ \\ 
        \addlinespace[0.25em]
        $C_{L^3}$ &  $27 \leq \sum_{i=11}^{15} \#Char(X_i) \leq 29 $ \\ 
        \addlinespace[0.25em]
         $C_{L^4}$& $82 \leq \sum_{i=1}^{15} \#Char(X_i) \leq 84$ \\
         \addlinespace[0.25em]
         \midrule 
         $C_{Meter}$& \multicolumn{1}{c}{Syllable Constraints} \\
         \midrule 
         $C_{M^1}$ &  $ \sum_{i=1}^{15} \#Syll(X_i) = 23$ \\
         \midrule
        $C_{Unary}$& \multicolumn{1}{c}{Unary Syllable and Character Constraints} \\
         & \multicolumn{1}{c}{First Line Constraints} \\
        \midrule 
        $C_{U^1}$ & $\#char(X_1) \leq 4 \bigwedge \#Syll(X_1) = 1$ \\
        $C_{U^2}$ & $\#char(X_2) \leq 6 \bigwedge \#Syll(X_2) = 2$ \\
        $C_{U^3}$ & $\#char(X_3) \leq 7 \bigwedge \#Syll(X_3) = 2$ \\
        
        \midrule
        & \multicolumn{1}{c}{Second Line Constraints} \\
        \midrule
        $C_{U^6}$ & $\#char(X_6) \leq 4 \bigwedge \#Syll(X_6) = 1$ \\
        $C_{U^7}$ & $\#char(X_7) = 10 \bigwedge \#Syll(X_7) = 4$ \\
        $C_{U^8}$& $X_8$ = "," \\
        $C_{U^9}$ & $ \#Syll(X_9) = 1$ \\
        $C_{U^{10}}$ & $ \#Syll(X_{10}) = 1$ \\
        $C_{U^{11}}$ & $ \#Syll(X_{11}) = 1$ \\
        \midrule
        & \multicolumn{1}{c}{Third Line Constraints} \\
        \midrule
        $C_{U^{12}}$ & $  \#Syll(X_{12}) = 2$ \\
        $C_{U^{13}}$ & $ \#Syll(X_{13}) = 2$ \\
        $C_{U^{14}}$ & $ 2 \leq \#Syll(X_{14}) \leq 3$ \\
        $C_{U^{15}}$ & $ 2 \leq \#Syll(X_{15}) \leq 3$ \\
        
         \bottomrule
         
         \end{tabular}
         }
   \caption{CSP model to tackle the RADNER-like sentences problem. It is a particular case of standardized sentences. Therefore, the various knapsack constraints are fixed. For instance, the total number of characters range is [82,84]. Additionally, several unary constraints are added to further model the RADNER problem. For instance, $C_{U^7}$: $\#char(X_7) = 10 \bigwedge \#Syll(X_7) = 4$, ensure that variables $X_7$ is assigned with a word of 10 letters and 4 syllables. Remark: The comma is viewed as a word, so the unary constraints associated with $X_8$ (i.e., $C_{U^8}$) enforce its assignment with a comma.}
   \label{RadnerCore:Model}
   \end{table}

\subsubsection{Language Dependent Rules, Here German}
\begin{itemize}
    \item (5) The first word of the first and of the second line was a word of 1 syllable and 3 letters.
    \item (6) The second word of the second line was a word of 3 syllables and 10 letters.
    \item (7) This word was followed by a relative clause that (8) began with 3 short words, each of 1 syllable.
    \item (9) In the first line there was one noun of 2 syllables and another word (not a noun) of 2 syllables.
    \item (10) The third line began with a word of 2 syllables, followed by (11) a noun of 2 syllables.
    \item (12) The next word was the verb of the sentence, composed of 3 syllables. 
    \item (13) The last word was a part of the verb and had 2 syllables.
\end{itemize}

Then, the RADNER test, designed initially in German, was adapted for other languages; while sharing the same core rules, several side constraints differ from one language to another. For instance, the first word of the first line must contain three characters and one syllable in the original version. Whereas in the Spanish version of the sentences (Radner-Vissum version \cite{radner-vissum:2008}), the first word satisfies the same rule or contains precisely two characters. Surprisingly, In the Portuguese version (Radner-Coimbra version \cite{radner-coimbra:2016}), the first word can also contain only one character.

The CP model described in Tab. \ref{RadnerCore:Model} defines the RADNER core problem independent from a particular target language. For a particular language, some constraints need to be hardened: For instance \[C_{U^1} :\#char(X_1) \leq 4 \bigwedge \#Syll(X_1) = 1, \] becomes in German, \[ C_{U^1} : \#char(X_1) = 3 \bigwedge \#Syll(X_1) = 1. \]  In the Spanish case it becomes, \[C_{U^1} : 2 \leq \#char(X_1) \leq 3 \bigwedge \#Syll(X_1) = 1.\] And finally in Portuguese, it is defined as, \[C_{U^1} : 1 \leq \#char(X_1) \leq 3 \bigwedge \#Syll(X_1) = 1. \]

From a CP point of view, $C_{U^1}$ (as described in the model) is a relaxation of the $C_{U^1}$ associated with a given language. Therefore, after the compilation of the RADNER CP model in $MDD_{final}$, \emph{CPTextGen} allows to perform a refinement of the solutions set, thanks to the MDD propagator or intersection as mentioned in Sec. \ref{sec:method}.
Thus, either the model or the solving or both are parametric.
\section{Results}
\label{sec:resultats}

\subsection{Experimental Conditions}

The approach described in Sec.~\ref{sec:method} was implemented in Java~17. The code is available upon request. % 
\paragraph{Application:}The \emph{CPTextGen} model is evaluated on a standardized RADNER-like sentence generation task. 
\paragraph{Generation:}The generation experiments were performed on a machine using an Intel(R) Xeon(R)~W-2175 CPU~@~2.50GHz with 256 GB of RAM and running under Ubuntu 18.04. 

\paragraph{Corpus Definition:}
The n-grams are extracted from Wikipedia English articles (multistream1-7)\footnote{https://dumps.wikimedia.org/enwiki/20240201/}. Then, the generation is performed with 4-grams. A substantially more significant corpus is needed to perform generation in 5-grams. In addition, if an n-gram does not belong to the NLTK English words list, the n-gram is filtered out. This is indeed a coarse lexicon criterion. But it avoids the introduction of bad n-grams (containing numbers, symbols, links). This criterion can be refined to adapt to more specific needs (e.g., language level).
\paragraph{Target Language:}
English RADNER-like sentences are generated. RADNER sentences in another language can be generated if another language is used as in corpus input.

\paragraph{Evaluation:} The sentence curation of the solution set is performed thanks to the PPL assessments of an LLM (GPT2).

\subsection{Results Analysis}
\begin{figure}
    \centering
    \resizebox{0.66\textwidth}{!}{ 
    \includesvg{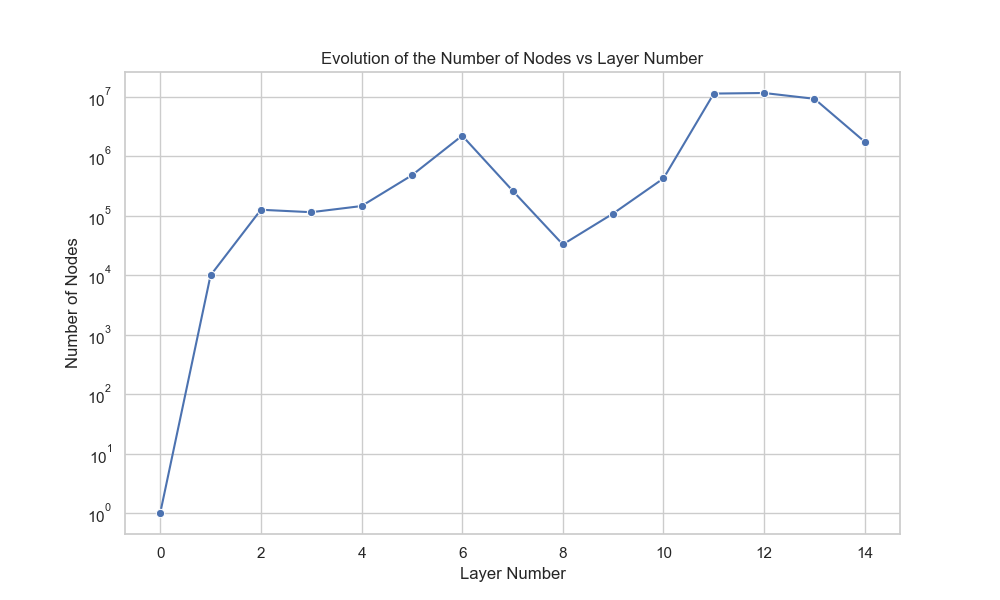}
    }
    \caption{Number of nodes for each layer of the MDD during the solving. The x-axis is the layer number, and the y-axis is the number of nodes in the layer (also called the width of the layer). Y-axis is in logarithmic scale. N.B.: There is at least one ingoing arc for each node in the MDD (except the root node). }
    \label{fig:evolution}
\end{figure}

Table \ref{tab:res} summarizes the benchmark performance on the Wikipedia articles dataset in 4-grams. The input corpus contains roughly 70 million of 4-grams. Only 6961 solutions are generated. For instance, with a corpus of only 3 million 5-grams, a similar approach generates seven thousand MNREAD sentences (see \cite{bonlarron-et-al:2023}). Thus, as expected, due to the various unary constraints and the meter constraints, the RADNER sentences problem is harder. Figure \ref{fig:evolution} further explains duration and memory consumption. The MDD width quickly reaches several dozen thousand nodes during the process. The impact of unary constraints that enforce a comma for the variables $X_8$ is dramatic (from 263469 nodes at layer 7 to 32738 nodes at layer 8).

In Table \ref{CP:sentences}, some of the best and worst-ranked sentences can be found. There is excellent discrimination between good and poor sentences thanks to PPL sorting for extreme values. Above 50 of PPL valuation sentences become poor semantically and then the PPL ranking is not helping so much. Therefore, going further above is not advisable. Finally, a bigger corpus is needed to ensure more variability in the generated sentences.
\begin{table}[tbp]
    \centering
    %\resizebox{0.5\textwidth}{!}{  
    \begin{tabular}{cccccc}\toprule
% [35:35:476] [230114mb] [INFORMATION] Done reducing					
%[35:35:729] [230114mb] [INFORMATION] #Solutions : 6961.0
%[35:35:824] [230114mb] [INFORMATION] #arcs: 3268
%[35:35:825] [230114mb] [INFORMATION] #nodes: 1682
       $n$  & nodes & arcs & mem (GB) & time (m) & sols   \\
       \midrule
       4  & 1682 & 3268 & 50 & 23 & 6961   \\
       
        \bottomrule
    \end{tabular}
    %}
    \caption{Number of arcs, nodes, solutions, gigabytes (GB), and minutes (m) for computing the MDD of RADNER-like sentences in 4-grams. 
    }
    \label{tab:res}
    \end{table}

\begin{table}[htbp]
    \centering
    \resizebox{!}{!}{  
    \begin{tabular}{ll}\toprule
    
      \multicolumn{2}{c}{RADNER-like generated sentences} \\ 
     \midrule
    sentences & PPL \\
    \midrule
    But after only three months of operations, they had been making very little progress & 28 \\
With only about three years of operations, the school is divided into seven sections & 31 \\
With only about three years of operations, the school is divided into seven boroughs & 32 \\
With only about three years of operations, they had been making very little progress & 32 \\
With only about three years of occupation, the school is divided into seven sections & 32 \\
With only about three years of operations, the state was divided into seven boroughs & 32 \\
With only about three years of retirement, he was ranked second among active players & 33 \\
With only about three years of occupation, the group was divided into seven sections & 33 \\
With only about three years of opposition, the house was divided into seven sections & 34 \\
With only about three years of occupation, the group has expanded into other markets & 34 \\
\midrule
His career over three years for conspiracy, and with the only other unlapped runners & 749 \\
His career over three years for conspiracy, to which the local area network standard & 750 \\ 
His career over three years for conspiracy, and with the station wagon body variants & 769 \\
The sagas describe three of the television, the left the country after Warsaw Chopin & 794 \\ 
The sagas describe three of the television, in the blood plasma within narrow limits & 856 \\ 
The sagas describe three of the television, and that the station wagon body variants & 862 \\ 
His career over three years for conspiracy, and that the only other unlapped runners & 862 \\ 
The sagas describe three of the television, the film was divided into Middle Harbour & 862 \\ 
His career over three years for conspiracy, and that the station wagon body variants & 1217 \\

         \bottomrule
         
    \end{tabular}
   }
   \caption{Radner-like sentences generated in 4-grams sorted according to PPL}
   \label{CP:sentences}
\end{table}

\section{Discussion}
\label{sec:discussion}

\subsection{Creative Settings}
\emph{CPTextGen} enables the integration of constraints related to the layout and combination of tokens, pre-tokens (words), or syllables. 
This makes it possible to directly model certain problems encountered in artistic context. For instance, when generating lyrics, constraints on sounds and rhymes could be defined and included (e.g., rhythmic templates \cite{Roy-Pachet:2013-meter}). Furthermore, the curation phase can be simplified (resulting in fewer outputs) by integrating LLM scoring earlier in the solving pipeline \cite{bonlarron-regin:2024}. This reduction in the number of outputs encourages the use of interactive mode, providing quick feedback to enhance creativity.

\subsection{The Challenge of Intertwining NLP and CP for Constrained Text Generation}

Faced with a constrained text generation problem, machine learning was not used due to the lack of available data, and no existing approach that proved appropriate from a combinatorial point of view was adopted. Instead, the problem of constrained text generation has been approached as a discrete combinatorial optimization problem. However, this paradigm choice raised a new challenge: how can certain constraints concerning the text itself, such as the meaning of a sentence, be expressed?

Why is this so difficult?
Let us take a step back; in NLP, the philosophy is to ``act as a human." Meanwhile, in CP, it is ``take a decision as a human". Satisfying rules requires \emph{decide as} while making sense rather requires \emph{act as}. 
Although \emph{act as} and \emph{decide as} seem close, they are very far apart conceptually and, in fact, when we observe the techniques used to achieve them. This ambivalence of point of view to generate constrained text is revealed in a loose combination of symbolic and statistical reasoning.
This naturally pushes us towards a symbolic approach to resolution (\emph{decide as} for the combinatorial aspect) but also forces us to look at more statistical considerations (\emph{act as} to express what we do not know how to model symbolically). 

\subsection{LLM Have the Attention but Not the Intention}

At its core, meaning is rooted in \emph{intention}. When we communicate, we have a specific purpose in mind, a message we want to convey. This intention drives the selection of words and the structure of our utterances. LLMs are attention-based. They process vast amounts of text data but do not have a personal ``I" behind these computations. 

If there is no \emph{I}, then there is no \emph{intention}. It is not easy to conceive the idea of meaning being a function of the \emph{intention} of a machine. And yet, by taking inspiration from bits of text written by humans (n-grams), do not we recover bits of \emph{intentions}? So we built sentences from bits of intention taken from corpora, and finally selected the best ones based on a democratic vote of general intention (distribution induced by training corpus) calculated by an LLM. The general intention does not necessarily make sense (because there is no \emph{I} behind a statistical distribution calculated from millions of texts either). However, at least it has the merit of being intelligible. The result is an illusion of meaning in the sentences selected.

\subsection{Almost Bona-Fide Sentences for LLM Evaluation?}

The proposed method enables the generation of ``unreasonably" constrained sentences, rendering them extremely unlikely to occur naturally. This suggests that the model consistently produces original sentences, effectively creating a new type of benchmark for evaluating LLMs.

A crucial question arises: if these sentences are so unlikely, does it imply that LLMs have never encountered them during their training phase? If so, could this approach serve as a new benchmark for assessing LLM performance? This question can be further explored by examining the perplexity (PPL) scoring of such a dataset. What should we make of a perfect sentence that receives a poor PPL score? Is it simply an expected outlier, like any other sample in the testing phase where the LLM assigns a poor PPL score to a perfectly valid sentence? Or could it be a subtle indication that while LLMs excel in well-structured scenarios driven by greedy decoding, they are susceptible to being misled by almost well-formed sentences? (\emph{With..., the school is divided into seven boroughs})

This observation raises a possibility: LLMs may excel at processing typical language patterns but struggle to effectively handle sentences that lie outside the boundaries of their training data. This suggests that their ability to generalize to unseen data may be limited, particularly when faced with highly constrained structures.

Investigating the PPL scores of sentences generated using this method could provide valuable insights into the strengths and limitations of LLMs in dealing with unexpected linguistic contexts. This approach could be further developed to create more sophisticated benchmarks that challenge LLMs to handle diverse and nuanced aspects of human language.

\section{Conclusion}
\label{sec:conclu}
The paper presents \emph{CPTextGen} framework as a hybrid modeling approach that loosely combines CP and NLP, to exploit the best of both worlds. Our comprehensive methodology guarantees the satisfaction of an arbitrary list of prerequisites while we use LLM expertise to select the best solutions.

\emph{CPTextGen} effectively solves a real-world problem, The RADNER sentences problem, while raising a fundamental question: How can CP be integrated with NLP? By combining symbolic and statistical approaches, we use CP to formulate and solve the well-defined part of the problem and LLM to deal with those we do not know how to solve and formulate clearly. In the particular case of sentence generation, as it is unclear how to define symbolically a meaning constraint, we rely on statistical computation to ensure the intelligibility of a sentence.

\section*{Acknowledgments}
We are grateful to Jerome Lebrun for proofreading and introducing us to the RADNER sentences, which played a key role in the development of this article.

This work has been supported by the French government, through the 3IA C\^ote d'Azur Investments in the
Future project managed by the National Research Agency (ANR) with the reference number ANR-19-P3IA-0002.

\bibliographystyle{unsrt}
\bibliography{ijcai24}

\end{document}